%% file: neurips_2025.tex
\documentclass{article}

 \usepackage[preprint]{neurips_2025}

\usepackage[utf8]{inputenc} 
\usepackage[T1]{fontenc}    
\usepackage{hyperref}       
\usepackage{url}            
\usepackage{booktabs}       
\usepackage{amsfonts}       
\usepackage{nicefrac}       
\usepackage{microtype}      
\usepackage{xcolor}         

\usepackage{graphicx} 
\usepackage{caption} 
\usepackage{algorithm}
\usepackage{algorithmic}

\usepackage{multirow}
\usepackage{arydshln}
\usepackage{array}
\usepackage{amsmath}
\usepackage{subcaption}
\usepackage{xspace}

\usepackage{enumitem}
\setlist[itemize]{leftmargin=*}

\title{LLM-based Human-like Traffic Simulation for Self-driving Tests}

\def\framework{\texttt{HDSim}\xspace}

\author{
  Wendi Li\textsuperscript{1} \quad
  Hao Wu\textsuperscript{1} \quad
  Han Gao\textsuperscript{1} \quad
  Bing Mao\textsuperscript{1} \quad
  Fengyuan Xu\textsuperscript{1,*} \quad
  Sheng Zhong\textsuperscript{1} \\
  \textsuperscript{1}National Key Lab for Novel Software Technology, Nanjing University \\
  \texttt{wendili@smail.nju.edu.cn, hao.wu@nju.edu.cn, gaohan@smail.nju.edu.cn} \\
  \texttt{\{maobing, fengyuan.xu, zhongsheng\}@nju.edu.cn} \\
}

\begin{document}

\maketitle

\input{0_abstract}

\input{1_intro}

\input{2_relatedwork}

\input{4_method}

\input{5_experiment}

\input{6_conclusion}

\bibliographystyle{plainnat}
\bibliography{neurips_2025}

\end{document}

%% file: 0_abstract.tex
\begin{abstract}

Ensuring realistic traffic dynamics is a prerequisite for simulation platforms to evaluate the reliability of self-driving systems before deployment in the real world. Because most road users are human drivers, reproducing their diverse behaviors within simulators is vital. Existing solutions, however, typically rely on either handcrafted heuristics or narrow data-driven models, which capture only fragments of real driving behaviors and offer limited driving style diversity and interpretability. To address this gap, we introduce \framework, an HD traffic generation framework that combines cognitive theory with large language model (LLM) assistance to produce scalable and realistic traffic scenarios within simulation platforms. The framework advances the state of the art in two ways: (i) it introduces a hierarchical driver model that represents diverse driving style traits, and (ii) it develops a Perception-Mediated Behavior Influence strategy, where LLMs guide perception to indirectly shape driver actions. Experiments reveal that embedding \framework into simulation improves detection of safety-critical failures in self-driving systems by up to 68\% and yields realism-consistent accident interpretability.

\end{abstract}

%% file: 1_intro.tex
\section{Introduction}

With end-to-end autonomous driving (E2E AD) systems nearing real-world deployment~\cite{tesla2025fsd}, robustness evaluation under realistic and interactive traffic scenarios has become increasingly critical. Simulation, unlike real-world testing, enables scalable and cost-effective exposure of AD systems to diverse safety-critical conditions~\cite{tan2024promptable}, and therefore constitutes a necessary stage prior to deployment. As a result, the integration of realistic background traffic is indispensable for simulation platforms such as CARLA~\cite{dosovitskiy2017carla}.
Among various traffic characteristics, the inclusion of human-driven (HD) vehicles is particularly important, as it introduces a distinctive source of behavioral variability. Traffic mixed with HD vehicles can expose a wider range of potential issues in tested AD systems compared to simplified, fully synthetic traffic. Figure~\ref{fig:introduction_fig} highlights the core advantage of incorporating HD traffic: its capacity to expose safety-critical weaknesses in E2E AD models that are often concealed in idealized environments, together with an illustrative example discovered by our \framework\ simulator.

However, existing background vehicle simulation methods fall short in accurately modeling human drivers. Most rely on rule-based or replay-based techniques~\cite{behrisch2011sumo, dosovitskiy2017carla, montali2023waymo, li2022metadrive}, which primarily produce homogeneous AD behaviors and fail to capture diverse driving behaviors. More recent approaches attempt to learn realistic driving behavior through data-driven strategies~\cite{zhang2025trajtok} and enhance behavioral diversity using prompt-driven learning methods~\cite{tan2024promptable}. Nevertheless, their reliance on real-world training data constrains the diversity of modeled driving styles, and they typically operate at the level of low-level control actions (e.g., acceleration, lane keeping). Such representations are \textit{insufficient to capture high-level style traits}, including aggressiveness, fatigue, and distraction, that influence not only actions but also perception, planning horizons, and risk sensitivity in complex traffic scenarios. Furthermore, they may raise privacy compliance concerns. It also remains unclear whether AD models trained on stylized datasets (e.g., aggressive driving data) might inadvertently degrade overall performance.

\input{introduction_fig}

This motivates a central \textit{research question}: How can we simulate interactive traffic environments populated with agents that demonstrate semantically grounded, temporally coherent, and perceptually integrated human driving styles, without relying on handcrafted rules or per-style retraining?  
We identify three key challenges in addressing this question:

(1) Lack of a formalized human driving style theory. Human driving styles emerge from latent factors encompassing persistent personality traits (e.g., aggressiveness) as well as temporary states (e.g., fatigue or distraction). Capturing these influences necessitates a structured  model that accounts for their hierarchical relationships and temporal evolution.

(2) Translating style into actionable behavior. High-level traits must be reliably embedded into driving decisions. This requires a generalizable mechanism that can integrate style into behavior in a context-aware and explainable way, ensuring safety and interpretability.

(3) Scalability and controllability in simulation. Achieving large-scale style diversity calls for a plug-and-play framework capable of dynamically activating traits across multiple agents, without the need for retraining or tightly coupled models.

To address these challenges, we introduce \framework, a simulation simulator focused on generating and modeling interactive driver agents with diverse, human-like driving styles, enabling realistic background traffic for AD testing. Our approach is grounded in the key cognitive insight that \textit{perception mediates behavior}, style traits influence how agents interpret the environment, which in turn shapes decision-making. Rather than injecting style at the action level, we model driving styles as perceptual biases that transform the agent's observations before control or planning.
Our \framework comprises three key points:

Model human driving styles. We develop a hierarchical human driver style model rooted in cognitive behavioral science. This model disentangles basic driving capabilities from three composable style influence layers: (i) \textit{personality influence} (e.g., aggressiveness), (ii) \textit{physiological influence} (e.g., fatigue), and (iii) \textit{attentional influence} (e.g., distraction), each associated with distinct temporal dynamics and influence patterns. This abstraction enables semantically rich style representation and modular composition of traits.

Generate stylized driver behaviors via LLM-guided perceptual transformation. To translate style traits into observable behaviors, we propose a novel Perception-Mediated Behavior Influence (PMBI) mechanism. Instead of directly modifying driving actions, PMBI transforms the agent's perception inputs using functional API programs synthesized by LLMs from natural language descriptions. These transformations simulate how cognitive traits subtly bias attention, perception, and environmental salience, resulting in realistic behavior without interfering with the AD model's control logic.

Enable scalable and flexible integration. Our design is non-intrusive and model-agnostic. \framework requires no retraining or architectural modifications of AD models, and seamlessly plugs into existing simulation platforms (e.g., CARLA). A runtime coordination layer schedules style activations across agents, enabling efficient multi-agent simulation with diverse, temporally evolving driving behaviors.

Extensive experiments show that our framework can reveal significant performance gaps in state-of-the-art AD models, uncovering up to 68\% more failure cases compared to standard testing setups, with strong alignment to real-world accidents. 
In addition, \framework can generate stylized traffic with high behavioral realism and support robust, scalable, and efficient simulation for large-scale deployment.

\noindent Our main contributions are as follows:
\begin{itemize}
\item We present the first framework for modeling diverse, stylized human driving behaviors in interactive traffic environments, substantially improving realism compared to existing rule-based or homogeneous simulation agents used for AD evaluation.
\item We introduce a hierarchical human driver style model that disentangles style-related influences from fundamental driving capabilities, and implement it using a heterogeneous AI approach to generate human-like driving behaviors.
\item We develop a simulation mechanism that is controllable, scalable, and computationally efficient, enabling seamless integration with existing AD simulators and straightforward adaptation to new driving styles without retraining.
\item Through extensive experiments on a state-of-the-art simulator, we demonstrate that incorporating stylized human drivers can reveal up to 68\% more hidden failures in AD models compared to conventional testing, while exhibiting strong consistency with real-world accident patterns.
\end{itemize}

\input{framework_fig}

%% file: introduction_fig.tex
\begin{figure}[tp]
    \centering
    \includegraphics[width=0.7\linewidth]{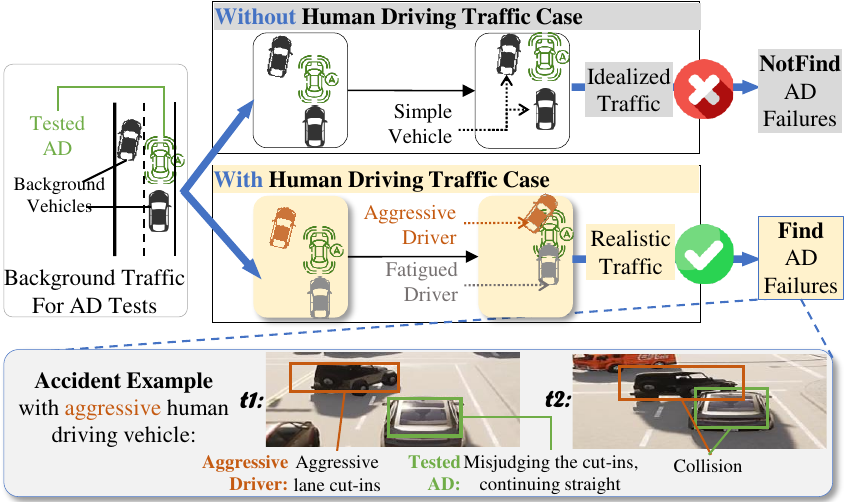}
    \caption{Diverse stylized traffic simulation reveals greater potential in identifying the shortcomings of AD models. Green box: tested AD, blue-white box: aggressive driver.}
    \label{fig:introduction_fig}
    \vspace{-10pt}
\end{figure}

%% file: framework_fig.tex
\begin{figure*}[htbp]
    \centering
    \includegraphics[width=0.95\linewidth]{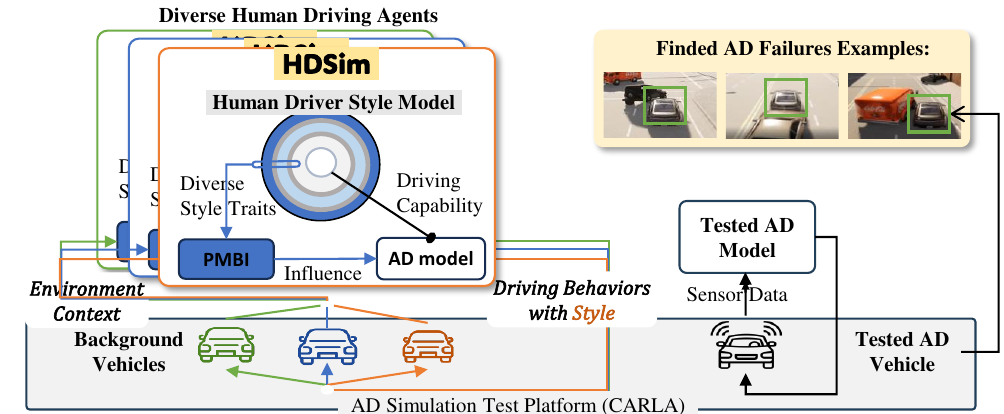}
    \caption{Overview of \framework: a human driver simulation framework for generating background traffic populated by driver agents with diverse human-like driving styles, each making decisions based on stylized subjective perception.}
    \label{fig:framework_fig}
\end{figure*}

%% file: 2_relatedwork.tex
\section{Related Work}
\label{sec:related}

\noindent\textbf{Driving Behavior Simulation in Traffic.}  
Existing traffic simulation approaches can be broadly classified into three categories: \textit{log-playback}, \textit{rule-based}, and \textit{learning-based}.  
Log-playback methods directly replay recorded real-world trajectories, ensuring realism in individual paths but lacking interactivity with other agents~\cite{montali2023waymo, li2022metadrive}, which limits their utility for testing adaptive decision-making.  
Rule-based approaches offer strong interpretability and controllability by explicitly encoding behavioral heuristics~\cite{behrisch2011sumo, dosovitskiy2017carla}, yet they typically yield homogeneous behaviors and fail to capture the variability present in real-world driving populations.  
Learning-based methods leverage imitation learning or reinforcement learning to improve behavioral realism~\cite{schulman2017proximal}, but their dependence on large-scale labeled datasets makes them less effective in rare, safety-critical, or long-tail scenarios~\cite{suo2021trafficsim, zhang2025trajtok}.  
More recently, LLM-powered simulation frameworks, such as ProSim~\cite{tan2024promptable}, enable customizable behaviors through fine-tuned models and natural-language instructions. While promising, these approaches mainly operate at the level of low-level vehicle control (e.g., throttle, brake, steering) and lack structured representations of high-level driver traits, which are essential for simulating style diversity in realistic traffic environments.

\noindent\textbf{Driver Style Simulation.}  
Approaches for simulating driver styles generally fall into two categories: \textit{parameter-based} and \textit{LLM-finetuned}.  
Parameter-based methods explicitly adjust a small set of behavioral parameters, for example desired speed, following distance, and lane-change aggressiveness, to mimic certain styles~\cite{dosovitskiy2017carla, niehaus1991expert}. These methods are computationally lightweight and interpretable but lack flexibility, scalability, and the ability to represent nuanced style combinations.  
LLM-finetuned methods~\cite{yang2024driving} generate style-specific driving behaviors by fine-tuning large language models on style-labeled datasets, allowing richer contextual decision-making. However, they often incur high computational overhead, face latency challenges in real-time simulation, and still struggle with stability.  
Both categories are generally restricted to a limited set of predefined styles (commonly ``aggressiv'' or ``cautious''), implemented in a rigid and fragmented fashion, which constrains their adaptability to complex and evolving traffic contexts.

%% file: 4_method.tex
\section{\framework Design}

This section first presents our cognitively inspired human driver style model, and then details how it is instantiated as a human driver simulator that can be non-intrusively integrated into existing AD testing environments such as CARLA.  
Based on these designs, \framework\ generates background traffic populated by diverse human driving agents, each equipped with style traits that influence their driving behavior, to test AD vehicles, as illustrated in Figure~\ref{fig:framework_fig}. This simulated traffic facilitates the identification of more safety-critical failures, as demonstrated by the examples in the Figure~\ref{fig:framework_fig}.

\subsection{Human Driver Style Model}

We propose a hierarchical model supported by knowledge from cognitive behavioral science. This model extracts cognitive factors, which lead to subtle driving action diversity, from basic driving capabilities and organizes them into a hierarchy. Such a hierarchical structure can accumulate the effects of multiple cognitive factors and guide LLMs to diversify abstracted human driving styles without knowing the concrete driving actions beforehand. 

Inspired by psychological theories of individual variability~\cite{elander1993behavioral, taubman2004multidimensional}, the model is structured as a set of concentric layers (Figure~\ref{fig:drivermodel_fig}), with an inner Driving Capability Layer (DCL) and three outer Style Influence Layers (SILs). This design supports modular composition of human-like driving behaviors in a scalable and semantically coherent way.

\input{drivermodel_fig}

At the center of this model, the DCL represents the driving action decisions made by human drivers according to their perceived in-situ contexts on the road, such as stopping at a red light, avoiding obstacles, and turning at a crossing. We assume this is shared by all rational human drivers with driver permits. Apparently, this part can be easily realized by current mainstream AD models. 

SILs, surrounding the DCL, encode those stackable driving effects of cognitive factors into three concentric rings, namely the personality influence layer, the physiological influence layer, and the attentional influence layer. 
The further a layer is from the center, the more transient and unstable its influence on the driving style is.

\begin{itemize}
    \item L1 -- Personality Influence Layer captures enduring psychological attributes, such as aggressiveness or cautiousness. These traits shape baseline risk perception and planning tendencies and remain stable over extended periods~\cite{TAUBMANBENARI2004323}.
    \item L2 -- Physiological Influence Layer models physiological states (e.g., fatigue, intoxication) that modulate behavioral control. We distinguish two types of influences: \textit{(i)} Incremental, such as fatigue, which accumulates gradually and triggers behavioral change once exceeding a cognitive threshold; and \textit{(ii)} 
    Episodic, such as alcohol impairment, which causes disruptions in judgment and reactivity~\cite{philip2001fatigue}.
    \item L3 -- Attentional Influence Layer accounts for attention dynamics in response to external complexity, grounded in Attention Resource Theory~\cite{kahneman1973attention}. Distracted driving, for instance, manifests as lagged awareness and missed hazards~\cite{endsley2017toward, schwarz2023distracted}.
\end{itemize}

Cognitive factors are categorized into these three layers based on their temporal stability and intensity, impacting the subtle driving action changes. A factor in the personality influence layer yields persistent driving characteristic changes, a factor in the physiological influence layer can be realized by periodical updates with a stylistic coherence mechanism, and a factor in the last layer captures unpredictable behavioral fluctuations, which can be realized as a stochastic process. When all factors in three layers are jointly modeled and coherently applied to concrete driving actions, the driving style is created for a simulated human driving vehicle. 


Thanks to this modeling, we can assign different human driving styles to each simulated human driving vehicle without the association of concrete driving actions made by the vehicle. Such a human driving style assignment can be in the form of a semantic description, which is an expertise of LLMs. Therefore, we let an LLM learn our human driving style model via in-context learning and generate a targeted driving style description at a high level for every simulated human driving vehicle. This description (
``An aggressive driver who relentlessly speeds through sparse traffic, consistently follows other vehicles at close distances, and disregards speed limits with reckless confidence.''
) will further be used to guide the micro-manipulation of every action of the assigned vehicle, achieving the targeted driving style influence, such as accelerating on a yellow light, hitting the brakes when the obstacle is far away, and turning by crossing solid lane marks.    

Please note that, in this stage, the LLM is only used once to generate the driving style description, which is used in a plug-and-play manner to indirectly influence the driving (details are provided in the next subsection). No training is required, and the application is transparent to the AD testing simulations, like CARLA. Thus, the proposed model makes our human driver simulator design scalable and non-intrusive to existing simulation platforms.

\input{stylized_behavior}

\subsection{Stylized Driver Behavior Generation}

For clarity, we first define the I/O of our stylized driver behavior generation. One input is a text description of the targeted human driving style $\mathcal{B}_{desc}$, which is provided by an LLM according to our human driving style model. $\mathcal{B}_{desc}$ is fixed during the simulation of style-assigned background vehicle. The other input is the context information \( \mathbf{ctx} \) at the current moment (i.e., the current step), which is provided by the AD testing simulations. \( \mathbf{ctx} \) contains all objects surrounding the simulated human driving vehicle (please note it is not the tested AD vehicle) and their physical states, like speeds and directions. On the other hand, the output of this stylized driver behavior generation is the driving action decision for the next step, which is denoted as \( \mathbf{D_t} \).

Our \framework design has two key components, as shown in Figure~\ref{fig:stylized_behavior_fig}. The first is an AD model taking as input the BEV (bird's eye view) images centered on it. \framework leverages this AD model to realize the DCL in our human driving style model. It supports our \framework vehicle to run from one place to another. Its driving actions, without the help of the second component, are denoted \( \mathbf{D^{dcl}} \). The second component is the PMBI mechanism powered by an LLM, which realizes the remaining three layers in our human driving style model. PMBI applies a targeted human driving style onto each \( \mathbf{D_t^{dcl}} \) in an indirect manner, which is done by manipulating the BEV image perceived by the first component. Such a manner is non-intrusive, so it does not impact the correctness of the used AD model, and it also provides good explainability for human inspectors when an accident occurs. 

Together, they can generate a sequence of driving actions \( \mathbf{D} \) with certain human driving characteristics, such as aggression, driving under the influence (DUI), and distraction. Additionally, \framework-controlled vehicles can interact with the tested AD vehicle on the road, making the simulation more realistic. Existing AD testing simulation platforms can easily replace rule-based or non-responsive background vehicles with \framework ones, without modifying the platforms. 

\subsection{Perception Mediated Behavior Influence (PMBI) Mechanism}




In this subsection, we elaborate on our PMBI mechanism, which is critical in \framework to naturally fuse the driving decision with the desired human driver style. Our design intuition has two points. First, we ask the AD model of our first component to take the lead in driving decisions, avoiding decision conflicts if there are multiple decision makers in the system. Second, we indirectly influence decisions of the AD model through the manipulation of what the model takes as input. In other words, we create an illusion gap between what is objective in the simulated context and what is subjective in the model's perception. This illusion gap will guide the AD model in adjusting its action decision a bit, which reflects the human driving style assigned to it. 

The creation of this illusion gap is automated by LLM with expert knowledge from cognitive behavioral science. We introduce a few principles and use them in the in-context learning for LLM. These principles teach the LLM how to interpret $\mathcal{B}_{desc}$ into a policy set \( \mathcal{P} = (p_1, p_2, p_3) \), where  \( p_1\), \( p_2 \), and \( p_3 \) instruct in text how the personality, physiological, and attentional factors make subtle behavior changes of driving actions, respectively.
For example,  an aggressive personality characteristic like ``confidently underestimates surrounding risks'' is interpreted into a $p_1$ policy of ``perceived distance of objects like vehicles in front is further than the real distance''. For another example, a physiological characteristic like ``DUI'' is interpreted into a $p_2$ policy of ``perceive straight lane marks as curved''.

Since our AD model perceives context information in the BEV form, we provide the LLM with a set of APIs (16 predefined perception modulation interfaces) along with documentation. These APIs are function calls designed to modify the contents of a BEV image, such as changing object size, changing object location, and changing traffic lights, with sensitivity coefficients across motion (e.g., speed), spatial (e.g., distance, size), temporal (e.g., traffic signals), and structural (e.g., lane alignment) dimensions~\cite{green1996usability}. We also provide a RAG with typical code examples, comprising 62 handcrafted scripts, to demonstrate how these APIs can be employed to implement a policy $p$. Examples are fed into the in-context learning way to LLM so that LLM can write high-quality code for the desired policy. Thus, we give the generic LLM the ability to translate text-based policies into API-based code instructions. 

During the simulation of a human-driven vehicle, policy translation is performed once for $p_1$ policies, periodically repeated with updated configurations for $p_2$ policies, and randomly triggered for $p_3$ policies.  
Temporal consistency across all action decisions is maintained by ensuring parameter consistency between two translations within the same policy layer, following the Weber–Fechner Law~\cite{fechner1860elemente} and theories of optic flow sensitivity~\cite{lee1976theory}.  
This approach also reduces the computational cost of LLM-based processing.

\noindent\textbf{Procedure Description.} At the moment of making $t$-th step driving action for one simulated vehicle, the procedure of PMBI is illustrated as follows. First, PMBI collects $ctx_t$ information from the AD testing simulation platform, like CARLA. For every object shown in the defined BEV area of the simulated vehicle, PMBI matches suitable API-based code instructions to change how it looks in the eye of our AD model. The consistency of change is also considered in our API implementations. After that, the changed BEV is passed to the AD model of this vehicle to generate $D_t$. The procedure of overall stylized driver behavior generation with PMBI as the Algorithm~\ref{alg:pmbi_alg}

To support robust and context-aware behavior induction, \framework further leverages a context-adaptive collaboration among LLM-based agents that produce behavior descriptions and scripts, reinforced by scene-scope semantic validation and runtime correction.

\input{pmbi_alg}

%% file: drivermodel_fig.tex
\begin{figure}[h]
    \centering
    \includegraphics[width=0.75\linewidth]{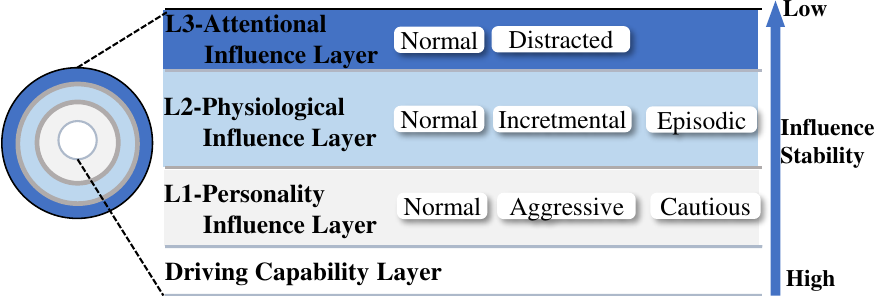}
    \caption{Hierarchical human driver style model.}
    \label{fig:drivermodel_fig}
\end{figure}

%% file: stylized_behavior.tex
\begin{figure}[h]
    \centering
    \includegraphics[width=0.8\linewidth]{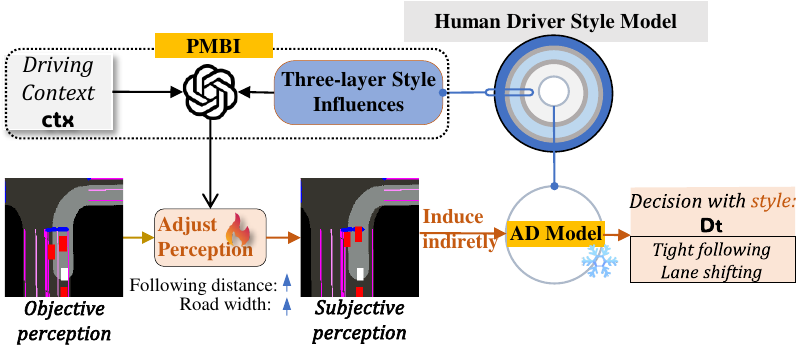}
    \caption{Stylized Driver Behavior Generation.}
    \label{fig:stylized_behavior_fig}
\end{figure}

%% file: pmbi_alg.tex
\begin{algorithm}[t]
\caption{Stylized Driver Behavior Generation with PMBI}
\label{alg:pmbi_alg}

\textbf{Input}: Style $s$, Simulator interface $\text{Sim}$, Simulation step $t$, Background AD model $\mathcal{M}_{AD}$, BEV input $\mathcal{X}_{\text{BEV}}$ \\
\textbf{Output}: Style-aligned driving decision $D_t$

\begin{algorithmic}[1]

\STATE $ctx_t \gets \text{Sim}.\text{GetContext}(t)$

\IF{$t==0$}
    \STATE $\mathcal{B}_{desc} \gets \text{LLM\_Generate}(s)$
    \STATE $\{p_1, p_2, p_3\} \gets \text{LLM\_Translate}(\texttt{Init}, \mathcal{B}_{desc})$
\ENDIF

\IF{$t$ is the update time of physiological factors}
    \STATE $p_2 \gets \text{LLM\_Translate}(\texttt{Update}, p_2)$
\ENDIF

\IF{$t$ is triggered randomly for attentional factors}
    \STATE $p_3 \gets \text{LLM\_Translate}(\texttt{ReInterpret}, \mathcal{B}_{desc})$ 
\ENDIF

\STATE $\mathcal{O}_t \gets \text{IdentifyObjects}(\mathcal{X}_{\text{BEV}})$, $\text{scripts}$=[]

\FOR{$o \in \mathcal{O}_t$}
        \STATE $APIs \gets \text{PMBI\_PolicyToAPIMapper}(o, \{p_1, p_2, p_3\})$
    \STATE compute adjustment parameters for $APIs$ based on last-time values to maintain consistency
    \STATE $\text{scripts} \gets [\text{scripts}, APIs ]$
\ENDFOR

\STATE $\mathcal{X'}_{\text{BEV}} \gets \text{adjust } \mathcal{X}_{\text{BEV}}$ by scripts

\STATE $D_t \gets \mathcal{M}_{AD}(\mathcal{X'}_{\text{BEV}})$
\STATE \textbf{return} $D_t$

\end{algorithmic}


\end{algorithm}

%% file: 5_experiment.tex
\section{Experiments}
\label{sec:experiment}



\subsection{Implementation}

All simulations are conducted in CARLA 0.9.10~\cite{dosovitskiy2017carla} within Town05, using 30 concurrent SimAgents per run. SimAgents are implemented using the Roach expert policy~\cite{zhang2021end}. To assess the impact of style influences, we validate representative traits across three style influence layers: L1 captures personality styles such as \textit{aggressive} and \textit{cautious}; L2 models physical traits with episodic (\textit{drunk}) and incremental (\textit{fatigued}) patterns; and L3 simulates transient attentional decline (\textit{distracted}). Each style is represented as a triplet (L1, L2, L3). L2 traits are updated every 2000 simulation steps, whereas L3 traits are triggered stochastically following a Poisson process with an arrival rate of 0.064. Each experiment contains 10 routes, repeated three times to ensure statistical robustness.
\subsection{Experimental Setup}

\noindent\textbf{Hardware.}
Rendering is performed on one NVIDIA RTX 4090 GPU. The simulation framework and multi-agent runtime are distributed across eight NVIDIA 2080 Ti GPUs. For language model inference, we use a hybrid configuration: LLaMA 3.1 is deployed locally on an A800 GPU for low-latency validation, while GPT-4o-mini is used for high-level behavioral reasoning and influence script generation.

\noindent\textbf{Metrics.} We adopt standard metrics from the CARLA Leaderboard v1~\cite{dosovitskiy2017carla}, including \textit{Driving Score} (DS) and \textit{Route Compliance} (RC), to evaluate critical driving safety over complete routes in AD testing.

\noindent\textbf{Baselines.} The conventional style-agnostic traffic simulation with a setting (normal, normal, normal) serves as the baseline test environment.

\noindent\textbf{Research Question.}
We evaluate the effectiveness, realism, and efficiency of \framework\ by addressing the three questions:

\textbf{RQ1.} Can \framework\ simulate diverse, style-rich human driving behaviors in traffic and effectively expose the weaknesses of AD models under these conditions?

\textbf{RQ2.} Do the AD failures identified by \framework\ correspond to real-world incidents, and how realistic are the generated stylized driving behaviors?

\textbf{RQ3.} Does \framework\ ensure robust stylized driving to support more reliable AD testing, while maintaining system scalability and efficiency for large-scale deployment?



\input{homogeneous_style_table}

\subsection{Exposing AD Weaknesses in Stylized Traffic}
\label{sec:style_aware_traffic}

To answer \textbf{RQ1}, we use \framework\ to construct three types of stylized traffic scenarios:  
\textit{style-homogeneous}, \textit{style-heterogeneous}, and \textit{selected challenging stylized} traffic.

In \textit{style-homogeneous} traffic (see Table~\ref{tab:homogeneous_style_table}), all agents in a scenario share the same driving style, defined by different combinations of SILs, including one, two, or all three layers.  
\textit{style-heterogeneous} traffic includes three styles, each involving one personality trait (normal, \textit{aggressive}, or \textit{cautious}), to represent a distinct type of human driver. These styles are further paired with their corresponding highest-risk outer-layer influences to ensure high-risk diversity and comprehensive style coverage. Each style controls 10 SimAgents in the simulation.
The \textit{selected challenging stylized} traffic (see Table~\ref{tab:ad_test_table}) set consists of three representative scenarios: the highest-risk single-SIL and highest-risk multi-SIL homogeneous settings, as well as the heterogeneous setting. These are used to evaluate AD model performance under particularly adverse conditions.

We first evaluate the effectiveness of stylized traffic using InterFuser~\cite{shao2023safety}, which achieves the strongest CARLA performance and serves as the representative AD model.  
To further explore the generalizability of \framework\ across model types, we test five additional state-of-the-art AD models:  
\textit{closed-loop models} include TFPP~\cite{jaeger2023hidden}, AIM~\cite{jaeger2023hidden};  
\textit{open-loop models} include VAD~\cite{jiang2023vad} and ST-P3~\cite{hu2022st};  
and the \textit{LLM-assisted model} is LMDrive~\cite{shao2024lmdrive}.

\noindent\textbf{Diverse Traffic Effectiveness in AD Tests.}
All stylized traffic scenarios result in significantly more hidden failures than the non-stylized baseline (Tables~\ref{tab:homogeneous_style_table}
), demonstrating that diverse human-style behaviors are effective in revealing AD weaknesses.
Within \textit{style-homogeneous} traffic (Table~\ref{tab:homogeneous_style_table}), the simulated traits and their combinations closely match both the framework design and observed real-world behavioral patterns. More stable traits yield consistent impacts: aggressive styles produce more critical failures than cautious ones, while fatigue leads to moderate effects. More unstable traits show greater variance across scenarios; among them, drunkenness leads to severe disruptions, whereas distraction mainly reduces routing completion. Co-occurring traits (e.g., aggressive and drunk) create strong synergistic effects that further amplify traffic risk.
Interestingly, \textit{style-heterogeneous} traffic induces only moderate effects (DS=$\downarrow$36.9\%, RC=87.3), likely due to offsetting interactions among conflicting styles, which lead to more balanced traffic dynamics.

\input{AD_test_table}
\input{caseStudy_fig_v2}

\noindent\textbf{Evaluating AD Models under Challenging Traffic.} As shown in Table~\ref{tab:ad_test_table}, the \textit{selected stylized challenging} traffic scenarios effectively stress-test all AD models, exposing failure modes often overlooked in conventional settings. Among them, closed-loop models achieve the highest overall performance but show the most limited adaptability under style perturbations. In contrast, open-loop models and the LLM-assisted model, despite their lower overall performance, maintain greater stability across diverse styles. This may be attributed to their broader consideration of interaction dynamics in model design or the benefits of LLM-guided reasoning.

\subsection{Realistic Stylized Accidents and Behaviors}
To answer \textbf{RQ2}, we investigate the realism of style-induced AD accident cases by aligning them with real-world reports. We further validate the similarity between simulated stylized behaviors and real-world driving styles, and compare our results against realism baselines.

\noindent\textbf{Style-Induced Failures: Real Case Analyses.}
We present three representative style-induced accident cases, each corresponding to a single style influence from L1 to L3. 
In Case 1, an aggressive driver performs a sudden cut-in without yielding, while the AD model fails to anticipate and continues straight, resulting in a collision (Figure~\ref{fig:CaseStudy_short-a}). In Case 2, a fatigued driver reacts late due to cognitive inertia and collides with an abruptly braking AD vehicle (Figure~\ref{fig:CaseStudy_short-b}). In Case 3, a distracted driver misses early cues due to degraded perception updates, and the AD model fails to brake in time, leading to a crash (Figure~\ref{fig:CaseStudy_short-c}).

The failure cases show strong alignment with real-world incidents (Caseid: 2005009501684, 2005041508481, 2007045403168) documented in NHTSA reports~\cite{nhtsa_nmvccs_viewer}, validating the realism of style-induced AD failures. This suggests that the latent failures identified in Tables~\ref{tab:homogeneous_style_table}–\ref{tab:ad_test_table} are not artifacts of simulation but likely to occur in real traffic, highlighting \texttt{HDSim}’s effectiveness in revealing critical AD blind spots without physical deployment.


\noindent\textbf{Validation of Simulated Style Realism.}
To validate style realism in \framework, five drivers (2–3 years' experience) annotated 3,901 style-unlabeled trajectories from the behavior-rich INTERACTION dataset~\cite{zhan2019interaction} into four single-SIL driving styles: (\textit{aggressive}, normal, normal), (\textit{cautious}, normal, normal), (normal, \textit{fatigued}, normal), and (normal, normal, \textit{distracted}). Three classifiers (RF, SVM, KNN) were then trained on these annotations to assess the alignment between simulated and real styles, with each style evaluated on 900 simulated trajectories.

\input{realism_table}

\input{realism_fig}

\noindent\textbf{Realism Baseline.}  
(1) CARLA’s parameter-based style module, supporting only (\textit{aggressive}, normal, normal) and (\textit{cautious}, normal, normal) styles;  
(2) ProSim, which simulates only concrete behavior instructions, not abstract styles. For comparison, we use \framework-generated stylized behavior descriptions as instruction inputs for ProSim.

As shown in Table~\ref{tab:style_eval}, \framework\ achieves up to 98.1\% classification F1-score, outperforming the baselines by an average of 23.3\% over CARLA and 21.3\% over ProSim. 
Figure~\ref{fig:all_featurewise_bar} further demonstrates that \framework\ exhibits minimal behavioral deviation from real-world driver styles (e.g., mean\_acc Wasserstein Distance~\cite{villani2008optimal}, $W=0.007$ for the (normal, \textit{fatigued}, normal)). Moreover, compared to ProSim, \framework\ achieves lower Wasserstein distances in speed ($\downarrow$38.16\%), acceleration ($\downarrow$27.42\%), and heading ($\downarrow$14.4\%), indicating superior realism.



\subsection{Driver Robustness and System Performance}

\noindent\textbf{Robust Decision-Making under Style Diversity.}  
For \textbf{RQ3}, we evaluate the performance of driver agents under five single-SIL driving styles within baseline traffic scenarios. As shown in Table~\ref{tab:robust_table}, each driver preserves robust driving performance while displaying distinct style traits, demonstrating that diverse style influences do not compromise driving robustness. This also rules out insufficient driver robustness as a confounding factor in the AD test results (Tables~\ref{tab:homogeneous_style_table}–\ref{tab:ad_test_table}).  
Notably, stylized drivers can achieve any driving capability (e.g., baseline performance) without model modification, enabled by \framework.

\input{robust_table}

\noindent\textbf{Evaluation of Simulation Efficiency and Scalability.}
For system performance in \textbf{RQ3}, runtime overhead remains minimal—CPU/GPU usage scales linearly with 30–70 driver agents, and the local LLM uses only 7.15~GB of GPU memory. LLM reasoning, triggered just 3–6 times per route, is latency-sensitive but fully parallelized. Overall runtime (with LLM inference) remains comparable to standard rule-based CARLA simulations. Full performance metrics are reported in Table~\ref{tab:memory_usage} (1 2080 Ti, 12~GB GPU).

\input{efficiency_table}

%% file: homogeneous_style_table.tex
\begin{table*}[t]
    \centering
    \caption{Performance of InterFuser AD Model in Style-Homogeneous Traffic. ``$\downarrow$'' indicates performance drop from the baseline (first row, separated by a horizontal line).}
    \label{tab:homogeneous_style_table}
    \setlength{\tabcolsep}{1.6mm}
    {\scriptsize
    \begin{tabular}{clllcc|lllcc}
   
    \toprule
    \noalign{
      \global\arrayrulewidth=0.8mm
    }
    & \multicolumn{3}{l}{\textbf{Style-Homogeneous Traffic}} & \multicolumn{2}{c|}{\textbf{Tested AD}} & \multicolumn{3}{l}{\textbf{Style-Homogeneous Traffic}} & \multicolumn{2}{c}{\textbf{Tested AD}} \\
    & L1 & L2 & L3 & DS & RC & L1 & L2 & L3 & DS & RC \\
    
    \midrule    
    \multirow{3}{*}{\textbf{One SIL influence:}} & normal & normal & normal & 100 & 100 & normal & \textit{drunk} & normal & $\downarrow$52.3\% & 61.2 \\       
    \noalign{
      \global\arrayrulewidth=0.1mm
    }
    \cline{2-6}
    \noalign{
      \global\arrayrulewidth=0.8mm
    }
    & \textit{aggressive} & normal & normal & \textbf{$\downarrow$54.5\%} & 100 & normal & \textit{fatigued} & normal & $\downarrow$38.1\% & 100 \\
    & \textit{cautious} & normal & normal & $\downarrow$21.9\% & 100 & normal & normal & \textit{distracted} & $\downarrow$35.4\% & 100 \\
        

    \noalign{
      \global\arrayrulewidth=0.1mm
    }
    \cline{1-11}
    \noalign{
      \global\arrayrulewidth=0.8mm
    }
    \multirow{4}{*}{\textbf{Two SIL influence:}} & normal & \textit{drunk} & \textit{distracted} & $\downarrow$31.3\% & 100 & normal & \textit{fatigued} & \textit{distracted} & $\downarrow$60.0\% & 89.0 \\ 
    & \textit{aggressive} & normal & \textit{distracted} & $\downarrow$62.7\% & 86.1 & \textit{cautious} & normal & \textit{distracted} & $\downarrow$48.3\% & 86.2 \\
    & \textit{aggressive} & \textit{drunk} & normal & \textbf{$\downarrow$64.7\%} & 100 & \textit{cautious} & \textit{drunk} & normal & $\downarrow$14.7\% & 100 \\
    & \textit{aggressive} & \textit{fatigued} & normal & $\downarrow$37.3\% & 100 & \textit{cautious} & \textit{fatigued} & normal & $\downarrow$37.0\% & 94.3 \\
    

    \noalign{
      \global\arrayrulewidth=0.1mm
    }
    \cline{1-11}
    \noalign{
      \global\arrayrulewidth=0.8mm
    }
    \multirow{2}{*}{\textbf{Three SIL influence:}} & \textit{aggressive} & \textit{drunk} & \textit{distracted} & \textbf{$\downarrow$57.0\%} & 100 & \textit{cautious} & \textit{drunk} & \textit{distracted} & $\downarrow$24.1\% & 100 \\
    & \textit{aggressive} & \textit{fatigued} & \textit{distracted} & $\downarrow$55.3\% & 100 & \textit{cautious} & \textit{fatigued} & \textit{distracted} & $\downarrow$46.7\% & 94.6 \\    
    \bottomrule
    \end{tabular}
    }

\end{table*}

%% file: AD_test_table.tex
\begin{table*}[t]
    \centering
    \caption{Performance of Diverse AD Models in Selected Challenging Stylized Traffic. Homogeneous (Homo.): one style controls 30 vehicles. Heterogeneous (Heter.): Three distinct styles control 10 vehicles each.}
    \label{tab:ad_test_table}
    \setlength{\arrayrulewidth}{0.2mm}
    \setlength{\tabcolsep}{1.4mm}
    {\scriptsize
    \begin{tabular}{llllcc|cc|cc|cc|cc}
        \toprule
         & \multicolumn{3}{l}{\textbf{Selected Challenging Stylized Traffic}} & \multicolumn{2}{c}{\textbf{TFPP}} & \multicolumn{2}{c}{\textbf{AIM}} & \multicolumn{2}{c}{\textbf{VAD}} & \multicolumn{2}{c}{\textbf{ST-P3}} & \multicolumn{2}{c}{\textbf{LMDrive}} \\
         & L1 & L2 & L3 & DS & RC & DS & RC & DS & RC & DS & RC & DS & RC \\
        \midrule
        \multirow{3}{*}{Homo.}& normal & normal & normal & 90.6 & 100 & 82.4 & 100 & 13.9 & 34.6 & 5.56 & 72.6 & 19.8 & 26.4 \\
         \noalign{
          \global\arrayrulewidth=0.1mm
        }
        \cline{2-14}
        \noalign{
          \global\arrayrulewidth=0.2mm
        }
         & \textit{aggressive} & normal &normal & $\downarrow$51.6\% & 94.8 & $\downarrow$67.6\% & 100 & $\downarrow$26.6\% & 30.5 & $\downarrow$37.1\% & 56.0 & $\downarrow$21.9\% & 22.3 \\
         & \textit{aggressive} & \textit{drunk} & normal & $\downarrow$54.5\% & 95.8 & $\downarrow$63.2\% & 100 & $\downarrow$33.1\% & 30.1 & $\downarrow$31.7\% & 55.7 & $\downarrow$23.3\% & 25.4 \\
         \noalign{
          \global\arrayrulewidth=0.1mm
        }
        \cline{1-14}
        \noalign{
          \global\arrayrulewidth=0.2mm
        }
          & normal & \textit{fatigued} & \textit{distracted} &  &  &  &  &  &  &  &  &  &  \\
         Heter. & \textit{aggressive} & \textit{drunk} & normal & $\downarrow$29.2\% & 100 & $\downarrow$24.5\% & 100 & $\downarrow$28.1\% & 32.0 & $\downarrow$25.9\% & 67.0 & $\downarrow$21.7\% & 25.6 \\
          & \textit{cautious} & normal & \textit{distracted} &  &  &  &  &  &  &  &  &  &  \\
        \bottomrule
    \end{tabular}
    }

\end{table*}

%% file: caseStudy_fig_v2.tex
\begin{figure*}[t]
  \centering
  \begin{subfigure}{0.48\linewidth}
    \includegraphics[width=\linewidth]{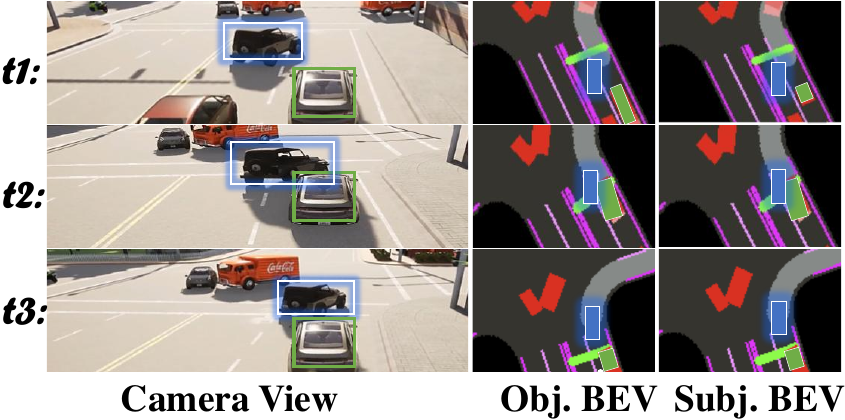}
    \caption{Case 1: (\textit{aggressive}, normal, normal)}
    \label{fig:CaseStudy_short-a}
  \end{subfigure}
  \hfill
  \begin{subfigure}{0.48\linewidth}
    \includegraphics[width=\linewidth]{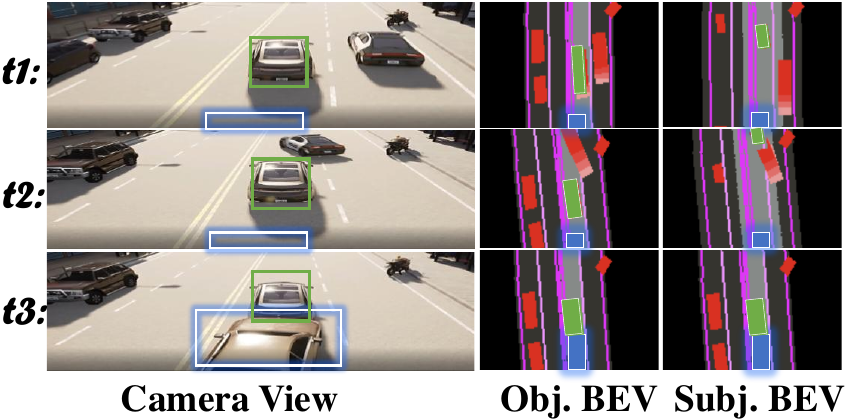}
    \caption{Case 2: (normal, \textit{fatigued}, normal)}
    \label{fig:CaseStudy_short-b}
  \end{subfigure}
  \hfill
  \begin{subfigure}{0.48\linewidth}
    \includegraphics[width=\linewidth]{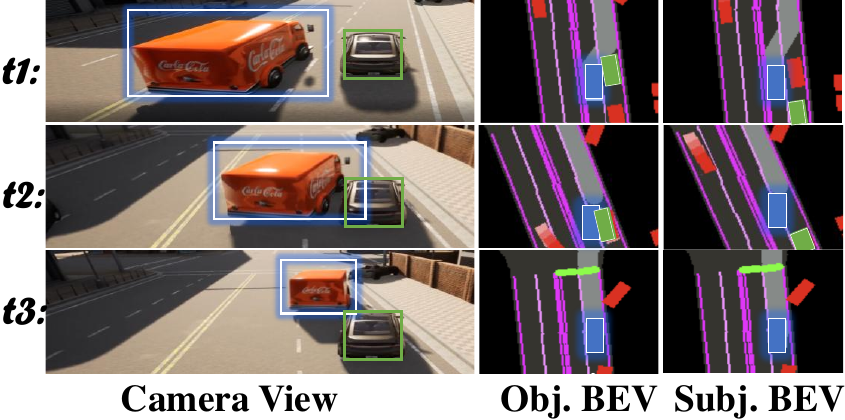}
    \caption{Case 3: (normal, normal, \textit{distracted})}
    \label{fig:CaseStudy_short-c}
  \end{subfigure}
  \caption{Case studies of AD failures involving stylized human drivers. Green boxes: tested AD vehicle, blue-white boxes: driver agents with specific styles, and red boxes: other items. Obj. BEV: Objective BEV; Subj. BEV: Subjective BEV.}
  \label{fig:caseStudy_fig}
\end{figure*}


%% file: realism_table.tex
\begin{table}[h]
    \centering
    \caption{F1-score (\%) of different driving styles evaluated using mean of RF, SVM, and KNN classifiers.}
    \label{tab:style_eval}
    \setlength{\arrayrulewidth}{0.2mm}
    {\footnotesize
    \begin{tabular}{lllc|c|c}
    \toprule
     L1 & L2 & L3 & CARLA & ProSim & \textbf{\framework(Ours)} \\
     \midrule
     \textit{aggressive} & normal & normal & 82.6 & 86.5 & \textbf{97.8} \\
     \textit{cautious} & normal & normal & 66.7 & 88.3 & \textbf{98.1} \\
     normal & \textit{distracted} & normal & -- & 0.00 & \textbf{45.0} \\
     normal & \textit{fatigued} & normal & -- & 53.5 & \textbf{72.7} \\
    \bottomrule
    \end{tabular}
    }

\vspace{-20pt}

\end{table}

%% file: realism_fig.tex
\begin{figure}[h]
  \centering
  \includegraphics[width=0.8\linewidth]{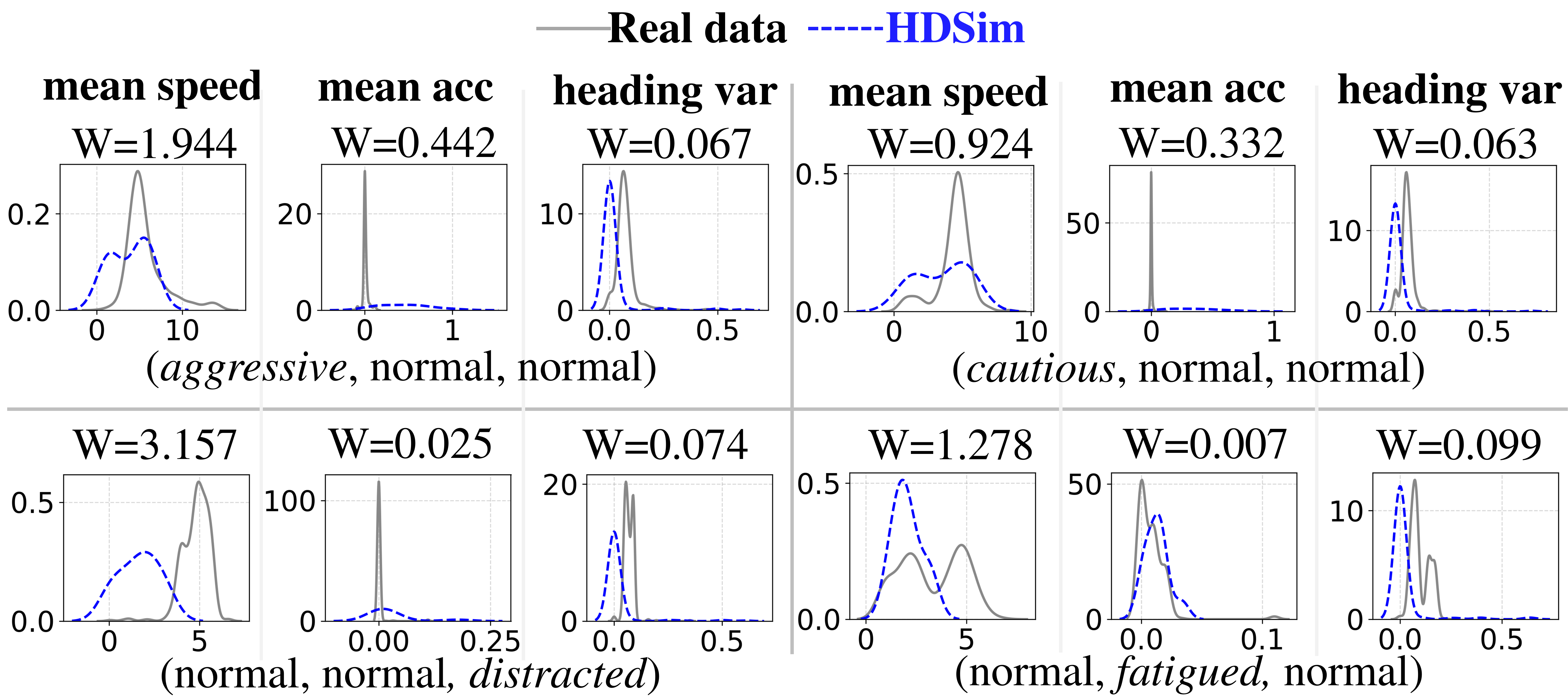}
  \caption{
    Real-world style alignment of \framework.
    Each subplot shows KDE curves comparing real vs simulated trajectories under different driver styles.
    X-axis denotes the value of each feature (e.g., speed, acceleration (acc)), and Y-axis shows the probability density.
    Smaller Wasserstein distance $W$ implies better alignment.
    }

  \label{fig:all_featurewise_bar}
\end{figure}

%% file: robust_table.tex
\begin{table}[h]
    \centering
    \caption{Robust driving evaluation under style diversity.}
    \label{tab:robust_table}
    \setlength{\tabcolsep}{5mm}
    {\footnotesize
    \begin{tabular}{lllcc}
    \toprule

    \multicolumn{3}{l}{\textbf{Tested Stylized Driver}} & \multicolumn{2}{c}{\textbf{Performanc}} \\
    L1 & L2 & L3 & DS & RC\\
    \midrule
    normal & normal & normal & 91.2 & 100 \\  
    \noalign{
      \global\arrayrulewidth=0.1mm
    }
    \cline{1-5}
    \noalign{
      \global\arrayrulewidth=0.8mm
    }
    \textit{aggressive} & normal & normal & $\downarrow$12.0 & 100 \\  
    \textit{cautious} & normal & normal & $\downarrow$3.20 & 100 \\
    normal & \textit{drunk} & normal & $\downarrow$12.6 & 100 \\  
    normal & \textit{fatigued} & normal & $\downarrow$8.80 & 100 \\  
    normal & normal & \textit{distracted} & $\downarrow$7.00 & 100 \\
    \bottomrule
    \end{tabular}
    }

\end{table}

%% file: efficiency_table.tex
\begin{table}[h]
    \centering
    \caption{System performance at different agent scales.}
    \label{tab:memory_usage}
    {\footnotesize
    \begin{tabular}{ccccc}
    \toprule
    \textbf{Agent} & \textbf{GPU} & \textbf{CPU} & \textbf{Driver Sim} & \textbf{Rule Sim} \\
    \textbf{Numbers} & \textbf{(GB/\%)} & \textbf{(GB/\%)} & \textbf{Step (s)} & \textbf{Step (s)} \\
    \midrule
    30 & 1.211 / 8\% & 20.4 / 1.3\% & 0.0229 & 0.0057 \\
    50 & 1.427 / 10\% & 22.4 / 1.3\% & 0.0341 & 0.0080 \\
    70 & 1.680 / 13\% & 24.0 / 1.3\% & 0.0480 & 0.0131 \\
    \bottomrule
    \end{tabular}
    }

\end{table}

%% file: 6_conclusion.tex
\section{Conclusion}

We present \framework, a cognitively inspired simulation framework for evaluating autonomous driving (AD) in interactive traffic with diverse human driving styles. Unlike existing rule-based or low-level behavior models, \framework models style as layered cognitive influences—personality, physiology, and attention—that bias perception and decision-making. A novel Perception-Mediated Behavior Influence (PMBI) mechanism uses LLM-generated programs to inject style-specific perceptual transformations without retraining or modifying AD models. Experimental results demonstrate its superiority over traditional tests in uncovering hidden issues in AD models by providing realistic human-driven traffic risks without physical deployment.


